\documentclass{article}

\usepackage[nohyperref, accepted]{icml2015} 

\usepackage{times}
\usepackage{epsfig}
\usepackage{graphicx}
\usepackage{amsmath}
\usepackage{amssymb}
\usepackage{subfigure}
\usepackage{capt-of}

\usepackage{natbib}
 
\icmltitlerunning{Online Tracking by Learning Discriminative Saliency Map with Convolutional Neural Network}

\begin{document}

\newcommand{\todo}[1]{#1}                
\newcommand{\ongo}[1]{#1}  
\newcommand{\gogo}[1]{{\color{magenta} #1}}  

\setlength\arraycolsep{2pt}

\newcommand{\x}{\mathbf{x}}
\newcommand{\y}{\mathbf{y}}
\newcommand{\w}{\mathbf{w}}
\newcommand{\z}{\mathbf{z}}
\newcommand{\h}{\mathbf{h}}
\newcommand{\M}{\mathbf{M}}
\newcommand{\R}{\mathcal{R}}
\newcommand{\Q}{\mathbf{Q}}
\newcommand{\ie}{i.e.}
\newcommand{\eg}{e.g.}

\twocolumn[
\icmltitle{Online Tracking by Learning Discriminative Saliency Map \\ with Convolutional Neural Network}

\icmlauthor{Seunghoon Hong\textsuperscript{$1$}}{maga33@postech.ac.kr}
\icmlauthor{Tackgeun You\textsuperscript{$1$}}{youtk@postech.ac.kr}
\icmlauthor{Suha Kwak\textsuperscript{$2$}}{suha.kwak@inria.fr}
\icmlauthor{Bohyung Han\textsuperscript{$1$}}{bhhan@postech.ac.kr}
\icmladdress{\textsuperscript{$1$}Dept. of Computer Science and Engineering, POSTECH, Pohang, Korea\\ \textsuperscript{$2$}INRIA--WILLOW Project, Paris, France}
\icmlkeywords{boring formatting information, machine learning, ICML}

\vskip 0.3in
]


\begin{abstract}
We propose an online visual tracking algorithm by learning discriminative saliency map using Convolutional Neural Network (CNN).
Given a CNN pre-trained on a large-scale image repository in offline, our algorithm takes outputs from \todo{hidden layers} of the network as feature descriptors since they show excellent representation performance in various general visual recognition problems.
The features are used to learn discriminative target appearance models using an online Support Vector Machine (SVM). 
In addition, we construct target-specific saliency map by back-propagating CNN features with guidance of the SVM, and obtain the final tracking result in each frame based on the appearance model generatively constructed with the saliency map.
Since the saliency map visualizes spatial configuration of target effectively, it improves target localization accuracy and enable us to achieve pixel-level target segmentation.
We verify the effectiveness of our tracking algorithm through extensive experiment on a challenging benchmark, where our method illustrates outstanding performance compared to the state-of-the-art tracking algorithms.%
\ongo{
} 
\end{abstract}


\begin{figure*}[!t]
\includegraphics[width=1\linewidth] {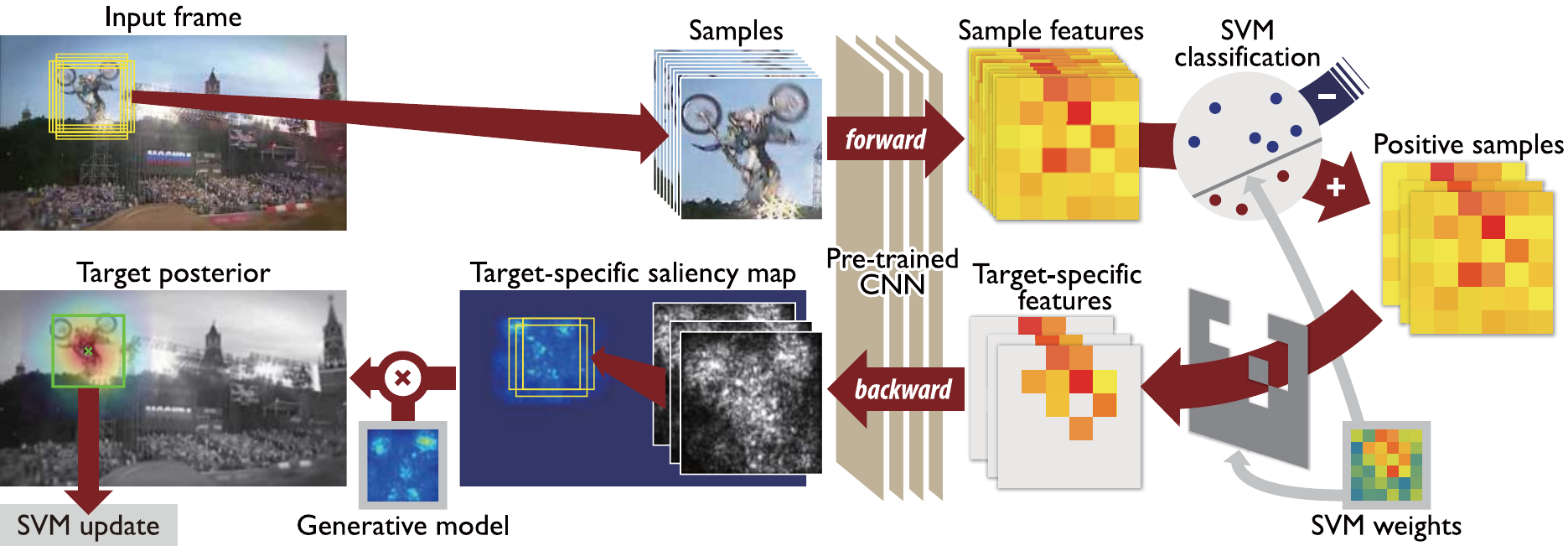}
\caption{Overall procedure of the proposed algorithm.
Our tracker exploits a pre-trained CNN for both image representation and target localization.
Given a set of samples on the input frame, we first extract their features using a pre-trained CNN (Section~\ref{sub:CNN_feature}), and classify them by the online SVM trained until the previous time step. 
For each positive sample, \ongo{we back-propagate the features relevant to target, which are identified by observing the model parameter of the SVM}, through the network to obtain a saliency map of the sample that highlights the regions discriminating target from background. 
The saliency maps of the positive examples are aggregated to build the target-specific saliency map  (Section~\ref{sub:saliency_map}).
Finally, tracking is performed by a sequential Bayesian filtering using the target-specific saliency map as observation.
To this end, a generative model is learned from target appearances in the previous saliency maps, and a dense likelihood map is calculated by convolution between the appearance model and the target-specific saliency map (Section~\ref{sub:generative_learning}).  
Based on the tracking result of the current frame, the SVM and generative model are updated for subsequent tracking (Section~\ref{sub:discriminative_learning}). 
}
\label{fig:overview}
\end{figure*}

\section{Introduction}
\label{sec:intro}
Object tracking has played important roles in a wide range of computer vision applications.
Although it has been studied extensively during past decades, object tracking is still a difficult problem due to many challenges in real world videos such as occlusion, pose variations, illumination changes, fast motion, and background clutter.
Success in object tracking relies heavily on how robust the representation of target appearance is against such challenges.

For this reason, reliable target appearance modeling problem has been investigated in recent tracking algorithms actively~\cite{bao:12,jia:12,mei:09,zhang:12,zhong:12,ross:04,han:08,babenko:11,hare:11,grabner:06,saffari:10}, which are classified into two major categories depending on learning strategies: generative and discriminative methods. 
In generative framework, the target appearance is typically described by a statistical model estimated from tracking results in previous frames.
To maintain the target appearance model, various approaches have been proposed including sparse representation~\cite{bao:12,jia:12,mei:09,zhang:12,zhong:12}, online density estimation~\cite{han:08}, incremental subspace learning~\cite{ross:04}, etc.
On the other hand, discriminative framework~\cite{babenko:11,hare:11,grabner:06,saffari:10} aims to learn a classifier that discriminates target from surrounding background. 
Various learning algorithms have been incorporated including online boosting~\cite{grabner:06, saffari:10}, multiple instance learning~\cite{babenko:11}, structured support vector machine~\cite{hare:11}, and online random forest~\cite{gall:11, schulter:11}.
These approaches are limited to using too simple and/or hand-crafted features for target representation, such as template, Haar-like features, histogram features and so on, which may not be effective to handle latent challenges imposed on video sequences.
Convolutional Neural Network (CNN) has recently drawn a lot of attention in computer vision community due to its representation power.
\cite{krizhevsky:12} trained a network using 1.2 million images for image classification and demonstrated significantly improved performance in ImageNet challenge~\cite{deng:12}.
Since the huge success of this work, CNN has been applied to representing images or objects in various computer vision tasks including object detection~\cite{girshick:14, sermanet:14, kaiming:14}, object recognition~\cite{oquab:14,donahue:14,zhang:14}, pose estimation~\cite{Toshev:14}, image segmentation~\cite{hariharan:14}, image stylization~\cite{karayev:14}, etc.

Despite such popularity, there are only few attempts to employ CNNs for visual tracking since offline classifiers are not appropriate for visual tracking conceptually and online learning based on CNN is not straightforward due to large network size and lack of training data.
In addition, the feature extraction from the deep structure may not be appropriate for visual tracking because the visual features extracted from top layers encode semantic information and exhibit relatively poor localization performance in general.
\cite{fan:10} presents a human tracking algorithm based on a network trained offline, but it needs to learn \ongo{a separate class-specific network} to track other kind of objects.
On the other hand, \cite{li:14} proposes a target-specific CNN for object tracking, where the CNN is trained incrementally during tracking with new examples obtained online. 
The network used in this work is shallow since learning a deep network using a limited number of training examples is challenging, and the algorithm fails to take advantage of rich information extracted from deep CNNs.
There is a tracking algorithm based on a pre-trained network~\cite{wang:13}, where a stacked denoising autoencoder is trained using a large number of images to learn generic image features.
Since this network is trained with tiny gray images and has no shared weight, its representation power is limited compared to recently proposed CNNs.

We propose a novel tracking algorithm based on a pre-trained CNN to represent target, where the network is trained originally for large-scale image classification.
On top of the \todo{hidden} layers in the CNN, we put an additional layer of an online Support Vector Machine (SVM) to learn a target appearance discriminatively against background.
The model learned by SVM is used to compute a target-specific saliency map by back-propagating the information relevant to target to input layer~\cite{simonyan:14}.
We exploit the target-specific saliency map to obtain generative target appearance models (filters) and perform tracking with understanding of spatial configuration of target.
%
The overview of our algorithm is illustrated in Figure~\ref{fig:overview}, and the contributions of this paper are summarized below:
\begin{itemize}
\item Although recent tracking methods based on CNN typically attempt to learn a network in an online manner~\cite{li:14}, our algorithm employs a pre-trained CNN to represent generic objects for tracking and achieves outstanding performance empirically.
\item We propose a technique to construct a target-specific saliency map by back-propagating only relevant features through CNN, which overcomes the limitation of the existing method to visualize saliency corresponding to the predefined classes only. This technique also enable us to obtain pixel-level target segmentation. 
\item We learn a simple target-specific appearance filter online and apply it to the saliency map; this strategy improves target localization performance even with shift-invariant property of CNN-based features.
\end{itemize}

The rest of this paper is organized as follows. 
We first describe the overall framework of our algorithm in Section~\ref{sec:overview} and the detailed methodology is discussed in Section~\ref{sec:proposed}.
The performance of our algorithm is presented in Section~\ref{sec:experiments}.

\section{Overview of Our Algorithm}
\label{sec:overview}

Our tracking algorithm employs a pre-trained CNN to represent target.
In each frame, it first draws samples for candidate bounding boxes near the target location in the previous frame, takes their image observations, and extracts feature descriptors for the samples using the pre-trained CNN.
We found out that the features from the CNN capture semantic information of target effectively and handle various geometric and photometric transformations successfully as reported in~\cite{oquab:14,karayev:14,donahue:14}.
However, it may lose some spatial information of the target due to pooling operations in CNN, which is not desirable for tracking since the spatial configuration is a useful cue for accurate target localization.

To fully exploit the representation power of CNN features while preserving spatial information of target, we adopt the target-specific saliency map as our observation for tracking, which is generated by back-propagating target-specific information of CNN features to input layer.
This technique is inspired by \cite{simonyan:14}, where class-specific saliency map is constructed by back-propagating the information corresponding to the identified label to visualize the region of interest.
Since target in visual tracking problem belongs to an arbitrary class and its label is unknown in advance, the model for target class is hard to be pre-trained.

Hence, we employ an online SVM, which discriminates target from background by learning target-specific information in the CNN features; the target-specific information learned by the online SVM can be regarded as label information in the context of~\cite{simonyan:14}.
The SVM classifies each sample, and we compute the saliency map for each positive example by back-propagating its CNN feature along the pre-trained CNN with guidance of the SVM till the input layer.
Each saliency map highlights regions discriminating target from background. 
The saliency maps of the positive examples are aggregated to build the target-specific saliency map.
The target-specific saliency map alleviates the limitation of CNN features for tracking by providing important spatial configuration of target.

Our tracking algorithm is then formulated as a sequential Bayesian filtering framework using the target-specific saliency map for observation in tracking.
A generative appearance model is constructed by accumulating target observations in target-specific saliency maps over time, which reveals meaningful spatial configuration of target such as shape and parts.
A dense likelihood map of each frame is computed efficiently by convolution between the target-specific saliency map and the generative appearance model.
The overall algorithm is illustrated in Figure~\ref{fig:overview}.
Our algorithm exploits the discriminative properties of online SVM, which helps generate target-specific saliency map.
In addition, we construct the generative appearance model from the saliency map and perform tracking through sequential Bayesian filtering.
This is a natural combination of discriminative and generative approaches, and we take the benefits from both frameworks.

\section{Proposed Algorithm}
\label{sec:proposed}
This section describes the comprehensive procedure of our tracking algorithm.
We first discuss the features obtained from pre-trained CNN.
The method to construct target-specific saliency map are presented in detail, and how the saliency map can be employed for constructing generative models and tracking object is described.
After that, we present online SVM technique employed to learn target appearance in a discriminative manner sequentially.

\subsection{Pre-Trained CNN for Feature Descriptor}
\label{sub:CNN_feature}

To represent target appearances, our tracking algorithm employs a CNN, which is pre-trained on a large number of images.
The pre-trained generic model is useful especially for online tracking since it is not straightforward to collect a sufficient number of training data.
In this paper, R-CNN~\cite{girshick:14} is adopted as the pre-trained model, but other CNN models can be used alternatively.
Out of the entire network structure, we take outputs from the first fully-connected layer as they tend to capture general characteristics of objects and have shown excellent generalization performance in many other domains as described in \cite{donahue:14}.

For a target proposal $\x_i$, the CNN takes its corresponding image observation $\z_i$ as its input, and returns an output from the first fully-connected layer $\phi(\x_i)$ as a feature vector of $\x_i$.
We apply the SVM to each CNN feature vector $\phi(\x_i)$ and classify $\x_i$ into either positive or negative.

\subsection{Target-Specific Saliency Map Estimation}
\label{sub:saliency_map}
%
For target tracking, we first compute SVM scores of candidate samples represented by the CNN features and classify them into target or background.
Based on this information, one na\"{i}ve option to complete tracking is to simply select the optimal sample with the maximum score as $$\x^* = \arg\max_i \w^{\top} \phi(\x_i).$$
However, this approach typically has the limitation of inaccurate target localization since, when calculating $\phi(\x_i)$, the spatial configuration of target may be lost by spatial pooling operations~\cite{fan:10}. 

To handle the localization issue while enjoying the effectiveness of CNN features, we propose the target-specific saliency map, which highlights discriminative target regions within the image.
This is motivated by the class-specific saliency map discussed in \cite{simonyan:14}. 
The class-specific saliency map of a given image $I$ is the gradient of class score $S_c(I)$ with respect to the image as
\begin{equation}
g_c(I) = \frac{\partial S_c(I)}{\partial I}.
\label{eq:gradient_class}
\end{equation}
The saliency map is constructed by back-propagation. 
Specifically, let $f^{(1)},\dots,f^{(L)}$ and $F^{(1)},\dots,F^{(L)}$ denote the transformation functions and their outputs in the network, where $F^{(l)}=f^{(l)}\circ f^{(l-1)}\circ\cdots\circ f^{(1)} (x)$ and $S_c(I) = F^{(L)}$.
Eq.~\eqref{eq:gradient_class} is computed using chain rule as
\begin{equation}
\frac{\partial S_c(I)}{\partial I} = \frac{\partial F^{(L)}}{\partial F^{(L-1)}}\frac{\partial F^{(L-1)}}{\partial F^{(L-2)}}\cdots \frac{\partial F^{(1)}}{\partial I}.
\label{eq:backprop_class}
\end{equation}
\ongo{
Intuitively, the pixels that are closely related to the class $c$ affect changes in $S_c$ more, which means that nearby regions of such pixels would have high values in saliency map.
}

When calculating such saliency map for object tracking, we impose target-specific information instead of class membership due to the reasons discussed in Section~\ref{sec:overview}. 
For the purpose, we adopt the SVM weight vector $\w = (w_1, \dots, w_n)^\top$, which is learned online to discriminate between target and background.
Since the last fully-connected layer corresponds to the online SVM, the outputs of the last two layers in our network are given by
%
\begin{eqnarray}
F^{(L)}&=&\w^TF^{(L-1)}+b \label{eq:svm_layer}\\
F^{(L-1)}&=&\phi(\x_i) \label{eq:feat_layer}.
\end{eqnarray}
%
Plugging Eq.~\eqref{eq:svm_layer} and \eqref{eq:feat_layer} into Eq.~\eqref{eq:backprop_class}, the gradient map of the target proposal $\x_i$ is given by
\begin{eqnarray}
g(\x_i) = \frac{\partial F^{(L)}}{\partial F^{(L-1)}}\frac{\partial F^{(L-1)}}{\partial \z_i} = \w^\mathrm{T} \left( \frac{\partial \phi(\x_i)}{\partial \z_i} \right),
\label{eq:gradient_target}
\end{eqnarray}
where $\z_i$ is the image observation of $\x_i$.

\todo{
Instead of using all entries in $\phi(\x_i)$ to generate target-specific saliency map, we only select the dimensions corresponding to positive weights in $\w$ since they have clearer contribution to make $\x_i$ positive. 
Note that every element in $\phi(\x_i)$ is positive due to ReLU operations in CNN learning.
Then, we obtain the target-specific feature $\phi^+(\x_i)$ as}
\begin{eqnarray}
\phi^+_k(\x_i) = \left\{
\begin{array}{cl}
w_k\phi_k(\x_i), & \text{if $w_k > 0$} \\
0, & \text{otherwise}
\end{array} 
\right., \nonumber
\end{eqnarray}
where $\phi_k(\x_i)$ denotes the $k$-th entry of $\phi(\x_i)$. 
Then the gradient of target-specific feature $\phi^+(\x_i)$ with respect to the image observation is obtained by
\begin{equation}
g(\x_i) = \frac{\partial \phi^+(\x_i)}{\partial \z_i},
\label{eq:gradient_target_specific}
\end{equation}
Since the gradient is computed only for the target-specific information $\phi^+(\x_i)$, pixels to distinguish the target from background would have high values in $g(\x_i)$.
%

\begin{figure}
\includegraphics[width=0.49\linewidth] {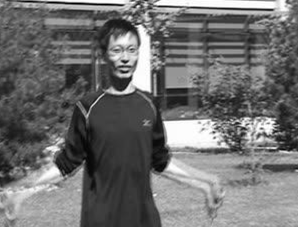}
\includegraphics[width=0.49\linewidth] {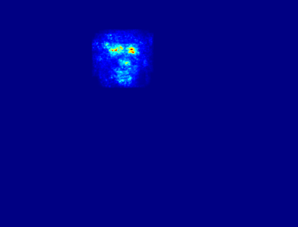}
\caption{An example of target-specific saliency map. The face of a person in left image is being tracked. The target-specific saliency map reveals meaningful spatial configuration of the target, such as eyes, a nose and lips.}
\label{fig:saliency_example}
\end{figure}

The target-specific saliency map $M$ is obtained by aggregating $g(\x_i)$ of samples with positive SVM scores in image space.
As $g(\x_i)$ is defined over sample observation $\z_i$, we first project it to image space and zero-pad outside of $\z_i$; we denote the result by $G_i$ afterwards. 
Then, the target-specific saliency map is obtained by taking the pixelwise maximum magnitude of the gradient maps $G_i$'s corresponding to positive examples, which is given by
\begin{equation}
M(p) = 
\max_{i} |G_i (p)|, ~~~\forall i \in \{ j | \w^\mathrm{T}\phi(\x_j) + b > 0 \},
\end{equation}
where $p$ denotes pixel location.
%
We suppress erroneous activations from background by considering only positive examples when aggregating sample gradient maps.
An example of target-specific saliency map is illustrated in Figure~\ref{fig:saliency_example}, where strong activations typically come from target areas and spatial layouts of target are exposed clearly.

\subsection{Target Localization with Saliency Map}
\label{sub:generative_learning}

Given the target-specific saliency map at frame $t$ denoted by $M_t$, the next step of our algorithm is to locate the target through sequential Bayesian filtering.
Let $\x_t$ and $M_t$ denote the state and observation variables at current frame $t$, respectively, where saliency map is used for measurement. 
The posterior of the target state $p(\x_t | M_{1:t})$ is given by
\begin{equation}
p(\x_t |  M_{1:t}) \propto p(M_t | \x_t) p(\x_t |  M_{1:t-1}),
\label{eq:bayesian}
\end{equation}
where $p(\x_t |  M_{1:t-1})$ denotes the prior distribution predicted from the previous time step, and $p(M_t | \x_t)$ means observation likelihood.

The prior distribution $p(\x_t |  M_{1:t-1})$ of target state at the current time step is estimated from the posterior at the previous frame through prediction, which is given by
\begin{equation}
\hspace{-0.165cm}p(\x_t |  M_{1:t-1}) = \int p(\x_t | \x_{t-1}) p(\x_{t-1} |  M_{1:t-1}) d\x_{t-1},
\label{eq:prediction}
\end{equation}
where $p(\x_t | \x_{t-1})$ denotes a state transition model.
\ongo{Target dynamics between two consecutive frames is given by a simple linear equation as}
\begin{equation}
\x_t = \x_{t-1} + \mathbf{d}_t + \boldsymbol{\varepsilon}_t,
\label{eq:dynamics}
\end{equation}
where $\mathbf{d}_t$ denotes a displacement of target location, and $\boldsymbol{\varepsilon}_t$ indicates a Gaussian noise.
Both $\mathbf{d}_t$ and $\boldsymbol{\varepsilon}_t$ are unknown before tracking in general, but is estimated from the samples classified as target by our online SVM in our case.
Specifically, $\mathbf{d}_t$ and $\boldsymbol{\varepsilon}_t$ are given respectively by
\begin{equation}
\mathbf{d}_t = \boldsymbol{\mu}_t  -\x_{t-1}^* , \qquad  \boldsymbol{\varepsilon}_t \sim \mathcal{N}(0, \Sigma_t),
\label{eq:motion}
\end{equation}
where $\x_{t-1}^*$ denotes the target location at the previous frame, and $\boldsymbol{\mu}_t$ and $\Sigma_t$ indicate mean and variance of locations of positive samples at the current frame, respectively.
From Eq.~\eqref{eq:dynamics} and \eqref{eq:motion}, the transition model for prediction is derived as follows:
\begin{equation}
p(\x_t | \x_{t-1}) = \mathcal{N}(\x_t - \x_{t-1}; \mathbf{d}_t, \Sigma_t).
\label{eq:process}
\end{equation}
Since the transition model is linear with Gaussian noise, computation of the prior in Eq.~\eqref{eq:prediction} can be performed efficiently by transforming the posterior $p(\x_{t-1} |  M_{1:t-1})$ at the previous step by $\mathbf{d}_t$ and applying Gaussian smoothing with covariance $\Sigma_t$.

The measurement density function $p(M_t | \x_t)$ represents the likelihood in the state space, which is typically obtained by computing the similarity between the appearance models of target and candidates.
In our case, we utilize $M_t$, target-specific saliency map at frame $t$, for observation to compute the likelihood of each target state. 
Note that pixel-wise intensity and its spatial configuration in the saliency map provide useful information for target localization.
%
%
At frame $t$, we construct the target appearance model $H_t$ given the previous saliency maps $M_{1:t-1}$ in a generative way.
Let $M_{k}(\x_{k}^*)$ denote the target filter at frame $k$, which is obtained by extracting the subregion in $M_{k}$ at the location corresponding to the optimal target bounding box given by $\x^*_{k}$.
The appearance model $H_{t}$ is constructed by aggregating the recent target filters as follows:
\begin{equation}
H_{t} = \frac{1}{m} \sum_{k=t-m}^{t-1} M_{k}(\x_{k}^*),
\label{eq:generative_model}
\end{equation}
where $m$ is a constant for the number of target filters to be used for model construction.
The main idea behind Eq.~\eqref{eq:generative_model} is that the local saliency map nearby the optimal target location in a frame plays a role as a filter to identify the target within the saliency map in the subsequent frames.
Since the target filter is computed based on $m$ recent filters, we need to store the $m$ filters to update the target filter.
Therefore, given the appearance model defined in Eq.~\eqref{eq:generative_model}, the observation likelihood $p(M_t| \x_t)$ is computed by simple convolution between $H_t$ and $M_t$ by
\begin{align}
p(M_t| \x_t) \propto H_t \otimes M_t (\x_t),
 \label{eq:saliency_score}
\end{align}
where $\otimes$ denotes convolution operator.
This is similar to the procedure in object detection, \eg, \cite{felzenszwalb:10}, where the filter is constructed from features to represent the object category and applied to the feature map to localize the object by convolution.

Given the prior in Eq.~\eqref{eq:prediction} and the likelihood in Eq.~\eqref{eq:saliency_score}, the target posterior at the current frame is computed simply by applying Eq.~\eqref{eq:bayesian}.
Once the target posterior is obtained, the optimal target state is given by solving the maximum a posteriori problem as
\begin{equation}
\x^*_t = \arg\max_\x p(\x_t |  M_{1:t}).
\end{equation}
Once tracking at frame $t$ is completed, we update the classifier based on $\x_t^*$, which is discussed next.

\subsection{Discriminative Model Update by Online SVM}
\label{sub:discriminative_learning}

\todo{
We employ an online SVM to learn a discriminative model of target.
Our SVM can be regarded as a fully-connected layer with a single node but provides a fast and exact solution in a single pass to learn a model incrementally.}

\todo{Given a set of samples with associated labels, $\{(\x'_i, y'_i)\}$, obtained from the current tracking results, we hope to update a weight vector $\w$ of SVM.}
The label $y'_i$ of a new example $\x'_i$ is given by
\begin{align}
y'_i = \left\{ 
\begin{array}{cc}
+1, &  ~~~\text{if }~~~ \x'_i = \x_t^* ~~~~~~~~~~~~~~~ \\
-1,  &  ~~~\text{if }~ \frac{\mathsf{BB}(\x^*_t) \cap \mathsf{BB}(\x'_i)}{\mathsf{BB}(\x^*_t) \cup \mathsf{BB}(\x'_i)} < \delta
\end{array}
\right.,
\label{eq:inc_gt}
\end{align}
%
where $\mathsf{BB}(\x)$ denotes the bounding box corresponding to the given state $\x$ and $\delta$ denotes a pre-defined threshold.
Note that the examples with the bounding box overlap ratios larger than $\delta$ are not included in the training set for our online learning to avoid drift problem.

Before discussing online SVM, we briefly review the optimization procedure of an offline learning algorithm.
Given training examples $\{(\x_i, y_i)\}$, the offline SVM learns a weight vector $\w  = (w_1, \dots, w_n)^\top$ by solving a quadratic convex optimization problem.
The dual form of SVM objective function is given by
\begin{equation}
\min_{0 \leq a_i \leq C} : W = \frac{1}{2} \sum_{i, j} a_i Q_{ij} a_j - \sum_{i} a_i + b \sum_i y_i a_i, \label{eq:svm_dual} \\
\end{equation}
where $\{a_i\}$ are Largrange multipliers, $b$ is bias, and $Q_{ij} = y_i y_j K(\x_i, \x_j)$.
In our tracking algorithm, the kernel function is defined by the inner product between two CNN features, \ie, $K(\x_i, \x_j) = \phi(\x_i)^\top \phi(\x_j)$.
In online tracking, it is not straightforward for conventional QP solvers to handle the optimization problem in Eq.~\eqref{eq:svm_dual} as training data are given sequentially, not at once.
Incremental SVM~\cite{diehl:03,cauwenberghs:01} is an algorithm designed to learn SVMs in such cases.
The key idea of the algorithm is to retain KKT conditions on all the existing examples while updating model with a new example, so that it guarantees an exact solution at each increment of dataset.
Specifically, KKT conditions are the first-order necessary conditions for the optimal solution of Eq.~\eqref{eq:svm_dual}, which are given by
\begin{eqnarray}
\frac{\partial W}{\partial a_i} & = & \sum_j Q_{ij} a_j + y_i b - 1 \left\{%
\begin{array}{rl}
  \geq 0,	& \textrm{if} \phantom{1} a_i = 0 		\\
  = 0, 		& \textrm{if} \phantom{1} 0 < a_i < C \phantom{10}	\\
  \leq 0,	& \textrm{if} \phantom{1} a_i = C,		\\
\end{array}%
\right. \label{eq:kkt_lagrangian} \\
\frac{\partial W}{\partial b} & = & \sum_j y_j a_j = 0, \label{eq:kkt_bias}
\end{eqnarray}
where $\frac{\partial W}{\partial a_i}$ is related to the margin of the $i$-th example that is denoted by $m_i$ afterwards.
By the conditions in Eq.~\eqref{eq:kkt_lagrangian}, each training example belongs to one of the following three categories: $E_1$ for support vectors lying on the margin ($m_i=0$), $E_2$ for support vectors inside the margin ($m_i < 0$), and $E_3$ for non-support vectors.


Given the $k$-th example, incremental SVM estimates its Lagrangian multiplier $a_k$ while retaining the KKT conditions on all the existing $k-1$ training examples.
In a nutshell, $a_k$ is initialized to 0 and updated by increasing its value over iterations.
In each iteration, the algorithm estimates the largest possible increment $\Delta a_k$ that guarantees KKT conditions on the existing examples, and updates $a_k$ and existing model parameters with $\Delta a_k$. 
This iterative procedure will stop when the $k$-th example becomes a support vector or at least one existing example changes its membership across $E_1$, $E_2$, and $E_3$.
We can generalize this online update procedure easily when multiple examples are provided as new training data.
With the new and updated Lagrangian multipliers, the weight vector $\w$ is given by
\begin{equation}
\w = \sum_{i \in E_1 \cup E_2} a_i y_i \phi(\x_i).
\end{equation}
For efficiency, we maintain only a fixed number of support vectors with smallest margins during tracking. 
We ask to refer to~\cite{diehl:03,cauwenberghs:01} for more details.
\todo{Also, note that any other methods for online SVM learning, such as LaSVM~\cite{bordes:05} and LaRank~\cite{bordes:07}, can also be adopted in our framework.}


\section{Experiments}
\label{sec:experiments}

This section describes our implementation details and experimental setting.
The effectiveness of our tracking algorithm is then demonstrated by quantitative and qualitative analysis on a large number of benchmark sequences.

\subsection{Implementation Details}
\label{sec:implementation}

For feature extraction, we adopt the R-CNN model built upon the Caffe library~\cite{jia:13}.
The CNN takes an image from sample bounding box, which is resized to $227\times 227$,
and outputs a 4096-dimensional vector from its first fully-connected ($\mathrm{fc}_6$) layer as a feature vector corresponding to the sample.
%
To generate target candidates in each frame, we draw $N(=120)$ samples from a normal distribution as $\x_i\sim\mathcal{N}(\x_{t-1}^*,\sqrt{wh}/2)$, where $w$ and $h$ denote the width and height of target, respectively.
The SVM classifier and the generative model are updated only if at least one example is classified as positive by the SVM. 
When generating training examples for our SVM, the threshold $\delta$ in Eq.~\eqref{eq:inc_gt} is set to $0.3$. 
The number of observations $m$ used to build generative model in Eq.~\eqref{eq:generative_model} is set to $30$.
To obtain segmentation mask, we employ {\em GrabCut}~\cite{rother:04}, where pixels that have saliency value larger than $70\%$ of maximum saliency are used as foreground seeds, and background pixels around the target bounding box up to 50 pixels margin are used as background seeds.
All parameters are fixed for all sequences throughout our experiment.


\subsection{Analysis of Generative Appearance Models}
\label{sec:analysis}

\begin{figure}[t]
\center
\includegraphics[width=1\linewidth] {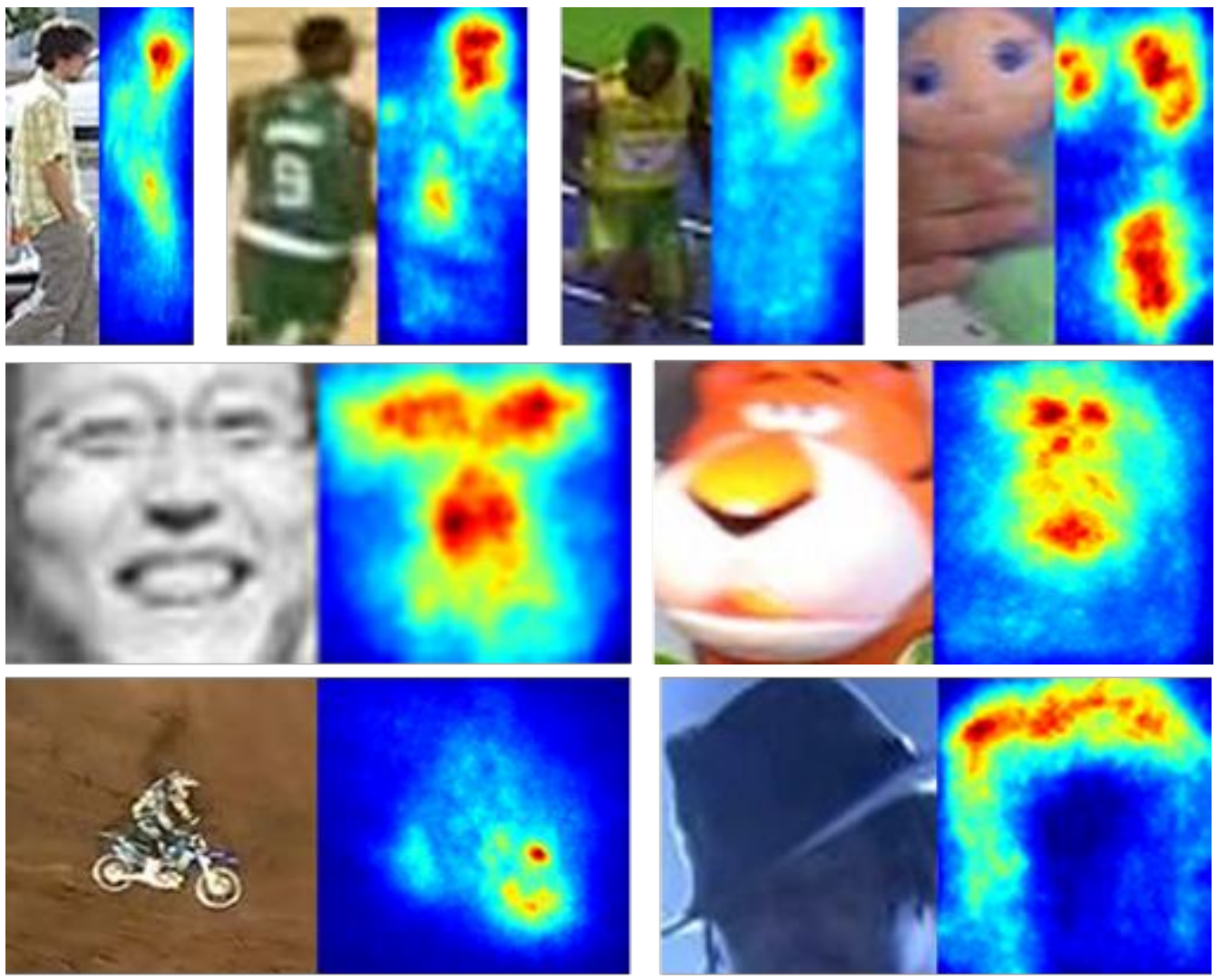}
\caption{Examples of generative models learned by our algorithm. In each example, the left and right image indicate the target and learned model, respectively. 
}
\label{fig:generative_models}
\vspace{-0.4cm}
\end{figure}

The generative model $H_t$ is used to localize the target using the target-specific saliency map. 
As described earlier, the target-specific saliency map shows high responses around discriminative target regions; our generative model exploits such property and is constructed using the saliency maps in the previous frames.
Figure \ref{fig:generative_models} illustrates examples of the learned generative models in several sequences. 
Generally, the model successfully captures parts and shape of an object, which are useful to discriminate the target from background.
More importantly, the distribution of responses within the model reveals the spatial configuration of the target, which provides a strong cue for precise localization.
This can be clearly observed in examples of face and doll, where the scores from the areas of eyes and nose can be used to localize the target.
When target is not rigid (\eg, person), we observe that the model has stronger responses on less deformable parts of the target (\eg, head) and localization relies more on the stable parts consequently.
\subsection{Evaluation}
\label{sec:quantitative}

%
%
\begin{figure}[t]
\centering
\subfigure{\includegraphics[width=0.8\linewidth]{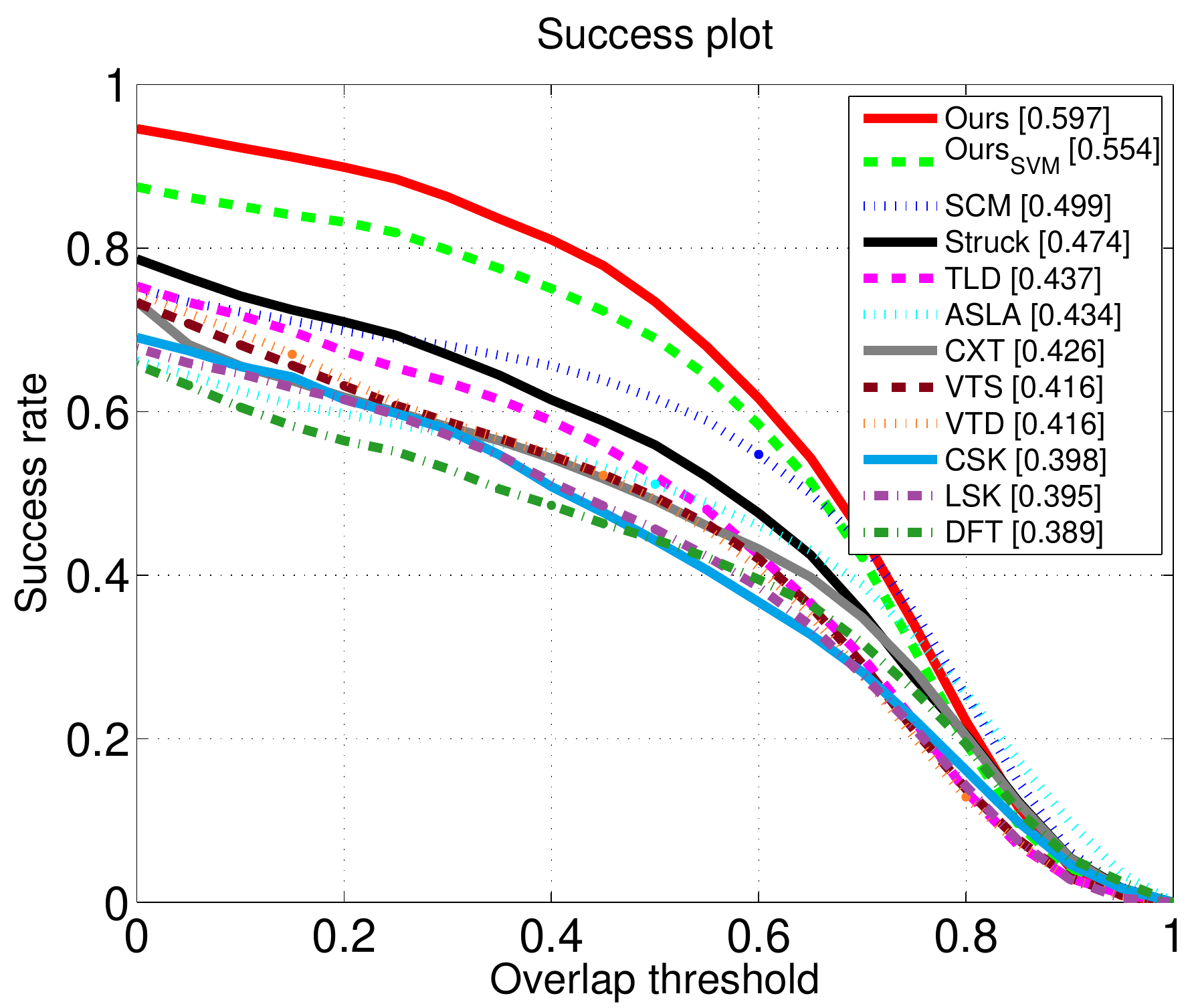}} \\
\subfigure{\includegraphics[width=0.8\linewidth] {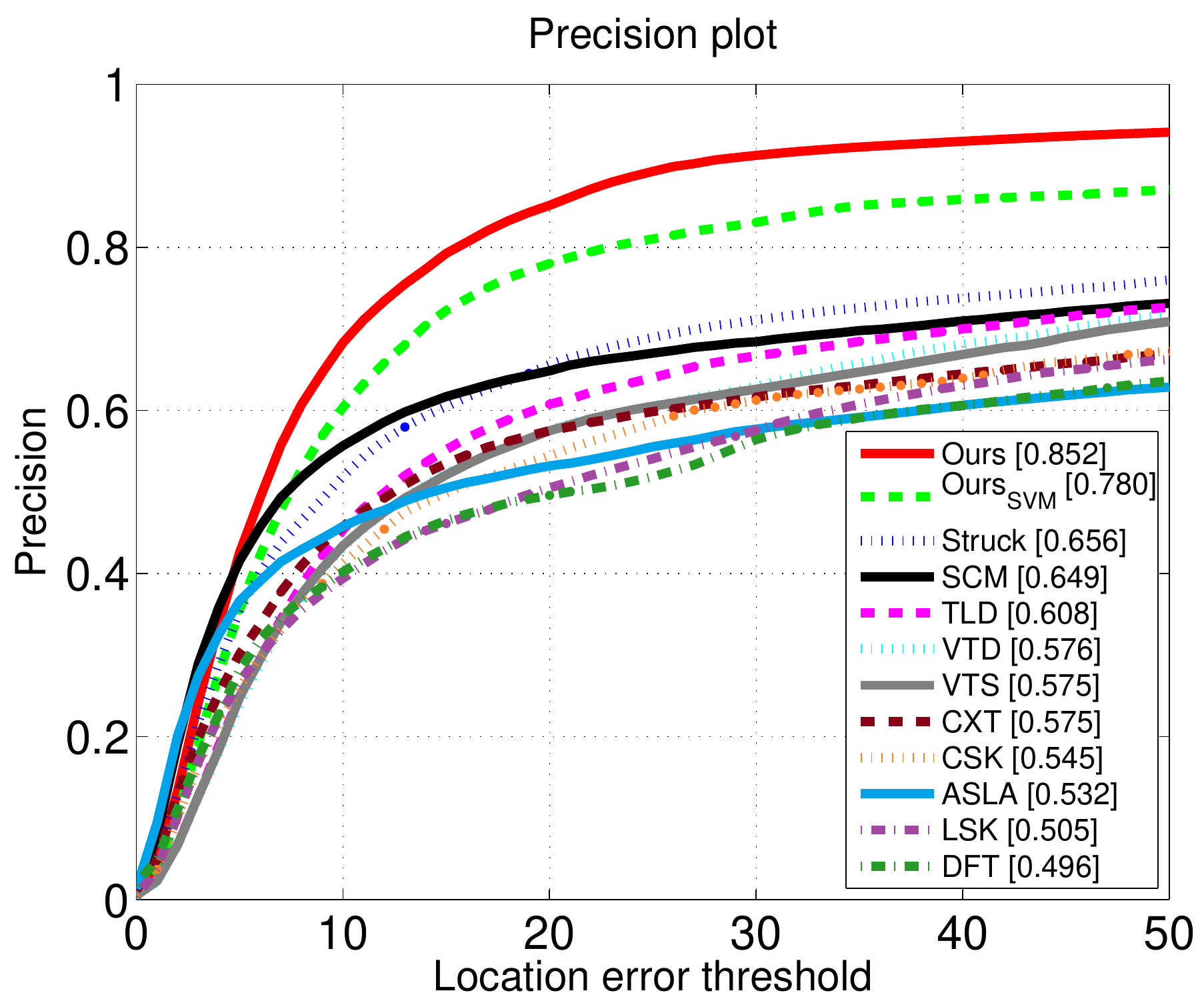}} 
\caption{Average success plot (top) and precision plot (bottom) over 50 benchmark sequences. Numbers in the legend indicate overall score of each tracker calculated by area under curve and distance at 20 pixels for success plot and precision plot. }
\label{fig:quantitative_benchmark_OPE}
\end{figure}
%

\paragraph{Dataset and compared algorithms}
To evaluate the performance, we employ all 50 sequences from the recently released tracking benchmark dataset~\cite{wu:13}. 
The sequences in the dataset involve various tracking challenges such as illumination variation, deformation, motion blur, background clutter, etc.
We compared our method with top 10 trackers in \cite{wu:13}, which include SCM~\cite{zhong:12}, Struck~\cite{hare:11}, TLD~\cite{kalal:12}, ASLA~\cite{jia:12}, CXT~\cite{thang:11}, VTD~\cite{kwon:09}, VTS~\cite{kwon:11}, CSK~\cite{henriques:12}, LSK~\cite{liu:11} and DFT~\cite{sevilla:12}. 
We used the reported results in~\cite{wu:13} for these tracking algorithms.

\paragraph{Evaluation methodology}
We follow the evaluation protocols in~\cite{wu:13}, where the performance of trackers are measured based on two different metrics: success rate and precision plots. 
In both metrics, the ratio of successfully tracked frames is measured by a set of thresholds, where bounding box overlap ratio and center location error are employed in success rate plot and precision plot, respectively.
We rank the tracking algorithms based on Area Under Curve (AUC) for success rate plot and center location error at 20 pixels for precision plot.

\begin{figure*}[!ht]
\begin{minipage}[c]{1\linewidth}
\scriptsize
\begin{center}
\captionof{table}{Average success rate scores on individual attributes. {\color{red}Red}: best, {\color{blue}blue}: second best.} \vspace{0.1cm}
\begin{tabular}{|c|c|c|c|c|c|c|c|c|c|c||c|c|}
\hline
&DFT	&LSK	&CSK	&VTS	&VTD	&CXT	&ASLA	&TLD	&Struck	&SCM	&$\text{Ours}_\text{SVM}$	&Ours	\\
\hline \hline
Illumination variation (25)	&0.383	&0.371	&0.369	&0.429	&0.420	&0.368	&0.429	&0.399	&0.428	&0.473	&\color{blue}{0.522}	&\color{red}{0.556}	\\
Out-of-plane rotation (39)	&0.387	&0.400	&0.386	&0.425	&0.434	&0.418	&0.422	&0.420	&0.432	&0.470	&\color{blue}{0.524}	&\color{red}{0.582}	\\
Scale variation (28)	&0.329	&0.373	&0.350	&0.400	&0.405	&0.389	&0.452	&0.421	&0.425	&\color{red}{0.518}	&0.456	&\color{blue}{0.513}	\\
Occlusion (29) 		&0.381	&0.409	&0.365	&0.398	&0.403	&0.372	&0.376	&0.402	&0.413	&0.487	&\color{blue}{0.539}	&\color{red}{0.563}	\\
Deformation (19)	&0.439	&0.377	&0.343	&0.368	&0.377	&0.324	&0.372	&0.378	&0.393	&0.448	&\color{blue}{0.623}	&\color{red}{0.640}	\\
Motion blur (12)	&0.333	&0.302	&0.305	&0.304	&0.309	&0.369	&0.258	&0.404	&0.433	&0.298	&\color{red}{0.572}	&\color{blue}{0.565}	\\
Fast motion (17)	&0.320	&0.328	&0.316	&0.300	&0.302	&0.388	&0.247	&0.417	&0.462	&0.296	&\color{red}{0.545}	&\color{red}{0.545}	\\
In-plane rotation (31)	&0.365	&0.411	&0.399	&0.416	&0.430	&0.452	&0.425	&0.416	&0.444	&0.458	&\color{blue}{0.501}	&\color{red}{0.571}	\\
Out of view (6)	&0.351	&0.430	&0.349	&0.443	&0.446	&0.427	&0.312	&0.457	&0.459	&0.361	&\color{red}{0.592}	&\color{blue}{0.571}	\\
Background clutter (21)	&0.407	&0.388	&0.421	&0.428	&0.425	&0.338	&0.408	&0.345	&0.458	&0.450	&\color{blue}{0.519}	&\color{red}{0.593}	\\
Low resolution (4)	&0.200	&0.235	&0.350	&0.168	&0.177	&0.312	&0.157	&0.309	&0.372	&0.279	&\color{blue}{0.438}	&\color{red}{0.461}	\\
\hline \hline
Weighted average	&0.389	&0.395	&0.398	&0.416	&0.416	&0.426	&0.434	&0.437	&0.474	&0.499	&\color{blue}{0.554}	&\color{red}{0.597}	\\
\hline
\end{tabular}
\label{tab:att_success}

\captionof{table}{Average precision scores on individual attributes. {\color{red}Red}: best, {\color{blue}blue}: second best.} \vspace{0.1cm}
\begin{tabular}{|c|c|c|c|c|c|c|c|c|c|c||c|c|}
\hline
&DFT	&LSK	&CSK	&VTS	&VTD	&CXT	&ASLA	&TLD	&Struck	&SCM	&$\text{Ours}_\text{SVM}$	&Ours	\\
\hline \hline
Illumination variation (25)	&0.475	&0.449	&0.481	&0.573	&0.557	&0.501	&0.517	&0.537	&0.558	&0.594	&\color{blue}{0.725}	&\color{red}{0.780}	\\
Out-of-plane rotation (39)	&0.497	&0.525	&0.540	&0.604	&0.620	&0.574	&0.518	&0.596	&0.597	&0.618	&\color{blue}{0.745}	&\color{red}{0.832}	\\
Scale variation (28)	&0.441	&0.480	&0.503	&0.582	&0.597	&0.550	&0.552	&0.606	&0.639	&0.672	&\color{blue}{0.679}	&\color{red}{0.827}	\\
Occlusion (29)	&0.481	&0.534	&0.500	&0.534	&0.545	&0.491	&0.460	&0.563	&0.564	&0.640	&\color{blue}{0.734}	&\color{red}{0.770}	\\
Deformation (19)	&0.537	&0.481	&0.476	&0.487	&0.501	&0.422	&0.445	&0.512	&0.521	&0.586	&\color{red}{0.870}	&\color{blue}{0.858}	\\
Motion blur (12)	&0.383	&0.324	&0.342	&0.375	&0.375	&0.509	&0.278	&0.518	&0.551	&0.339	&\color{red}{0.764}	&\color{blue}{0.745}	\\
Fast motion (17)	&0.373	&0.375	&0.381	&0.353	&0.352	&0.515	&0.253	&0.551	&0.604	&0.333	&\color{red}{0.735}	&\color{blue}{0.723}	\\
In-plane rotation (31)	&0.469	&0.534	&0.547	&0.579	&0.599	&0.610	&0.511	&0.584	&0.617	&0.597	&\color{blue}{0.720}	&\color{red}{0.836}	\\
Out of view (6)	&0.391	&0.515	&0.379	&0.455	&0.462	&0.510	&0.333	&0.576	&0.539	&0.429	&\color{red}{0.744}	&\color{blue}{0.687}	\\
Background clutter (21)	&0.507	&0.504	&0.585	&0.578	&0.571	&0.443	&0.496	&0.428	&0.585	&0.578	&\color{blue}{0.716}	&\color{red}{0.789}	\\
Low resolution (4)	&0.211	&0.304	&0.411	&0.187	&0.168	&0.371	&0.156	&0.349	&\color{blue}{0.545}	&0.305	&0.536	&\color{red}{0.705}	\\
\hline \hline
Weighted average	&0.496	&0.505	&0.545	&0.575	&0.576	&0.575	&0.532	&0.608	&0.656	&0.649	&\color{blue}{0.780}	&\color{red}{0.852}	\\
\hline
\end{tabular}
\label{tab:att_precision}
\end{center}
\vspace{0.1cm}
%

\center
\includegraphics[width=0.7\linewidth] {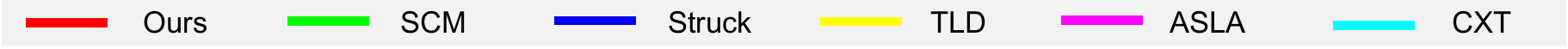}\\
\vspace{0.02cm}
\includegraphics[height=0.0798\textheight] {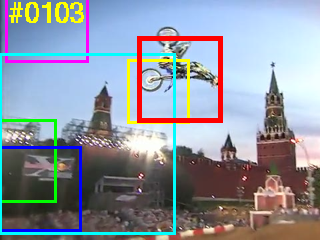}
\includegraphics[height=0.0798\textheight] {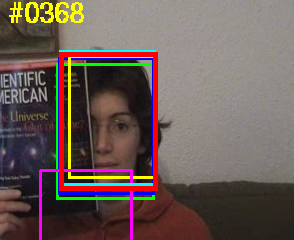}
\includegraphics[height=0.0798\textheight] {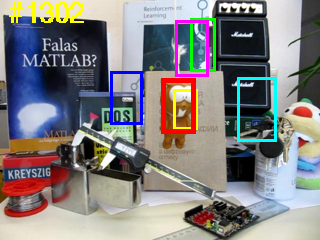}
\includegraphics[height=0.0798\textheight] {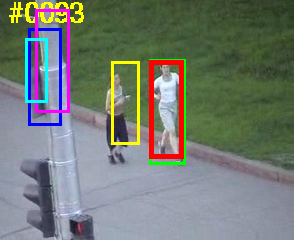}
\includegraphics[height=0.0798\textheight] {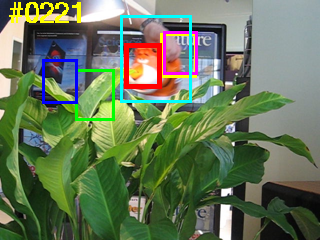}
\includegraphics[height=0.0798\textheight] {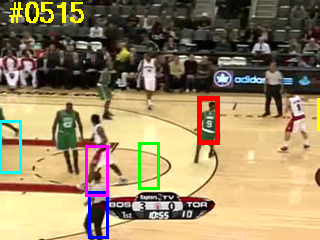}
\includegraphics[height=0.0798\textheight] {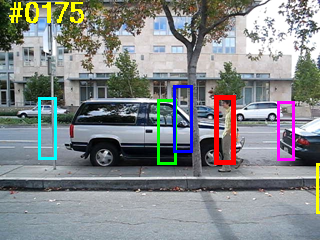}\\
\vspace{0.02cm}
\includegraphics[height=0.0798\textheight] {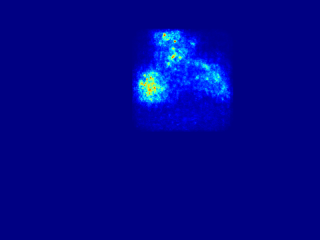}
\includegraphics[height=0.0798\textheight] {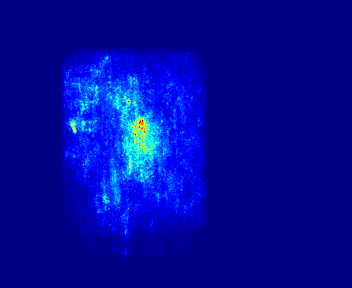}
\includegraphics[height=0.0798\textheight] {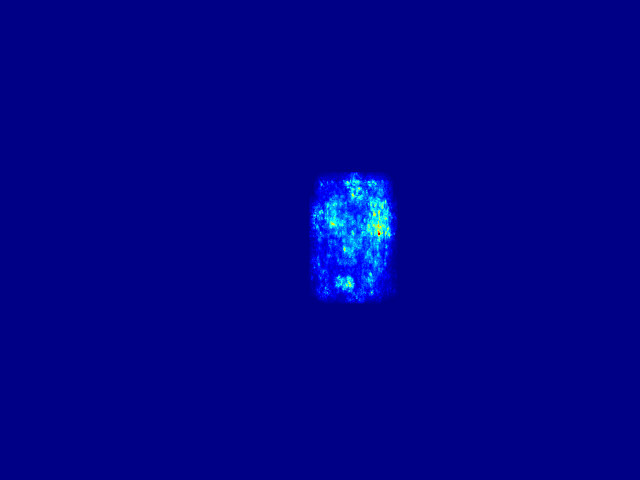}
\includegraphics[height=0.0798\textheight] {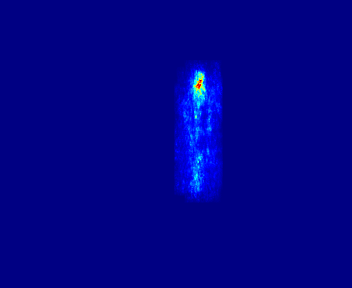}
\includegraphics[height=0.0798\textheight] {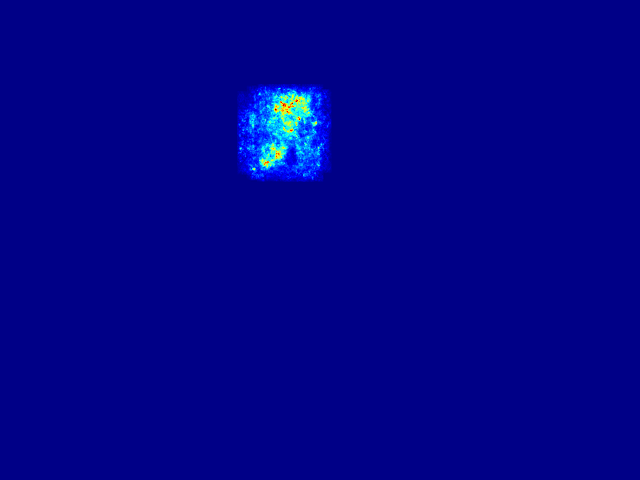}
\includegraphics[height=0.0798\textheight] {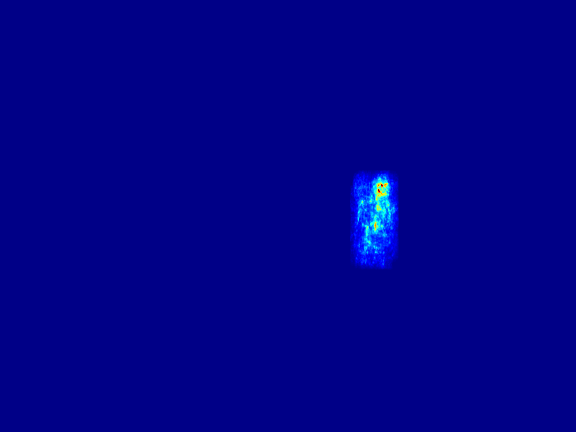}
\includegraphics[height=0.0798\textheight] {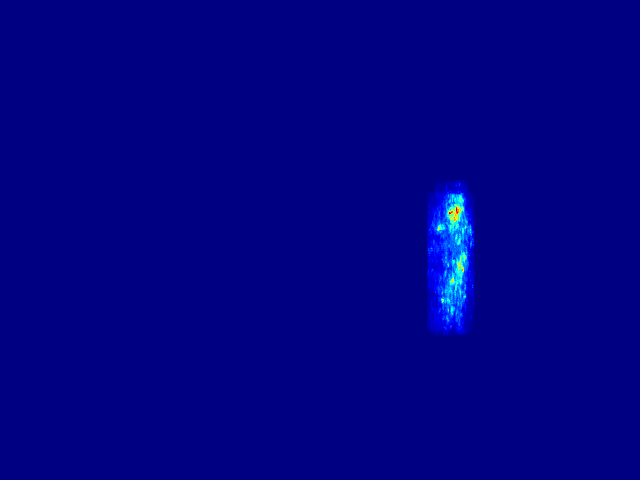}\\
\vspace{0.02cm}
\includegraphics[height=0.0798\textheight] {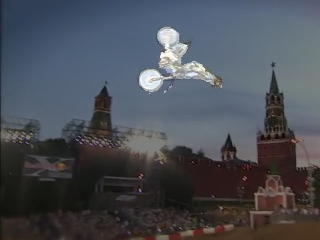}
\includegraphics[height=0.0798\textheight] {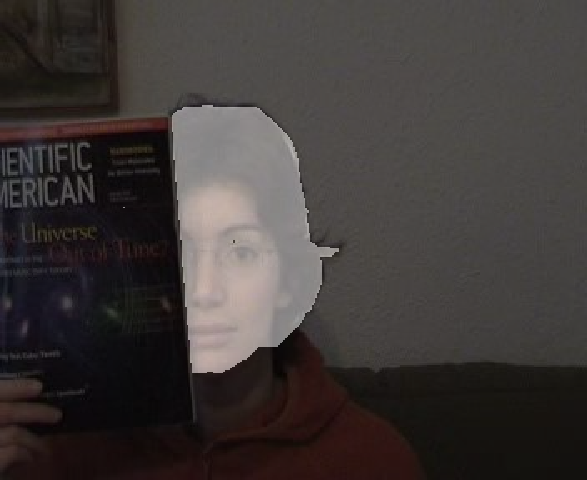}
\includegraphics[height=0.0798\textheight] {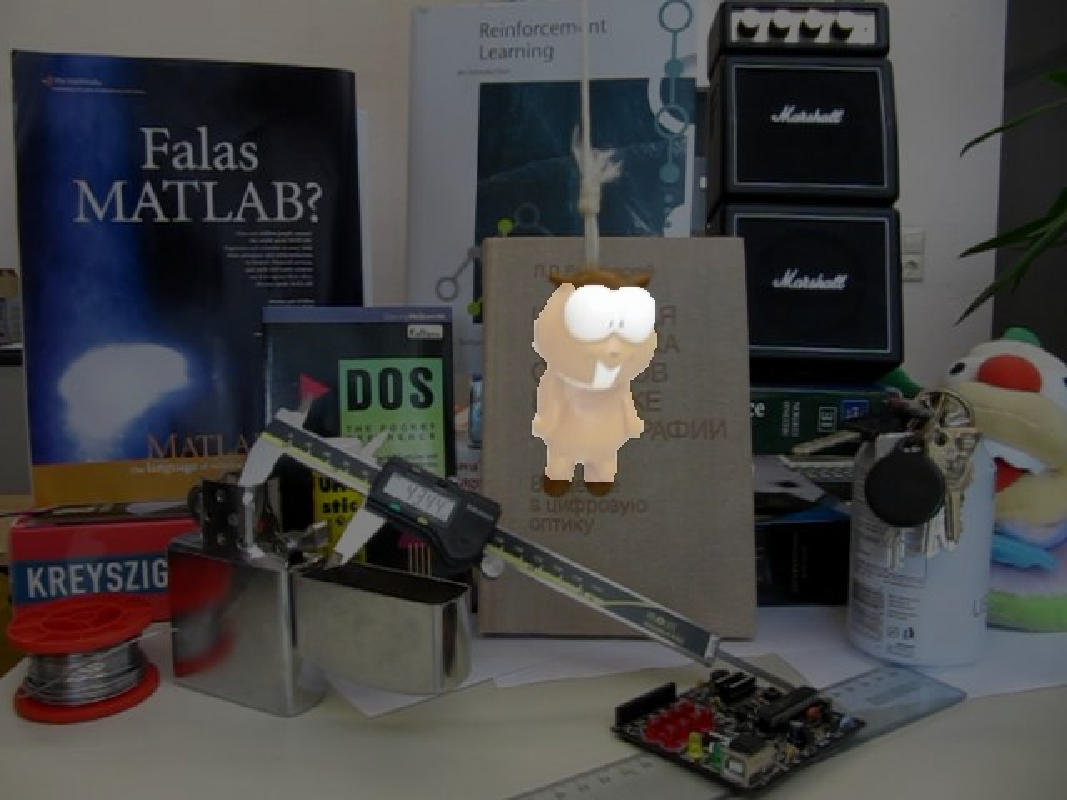}
\includegraphics[height=0.0798\textheight] {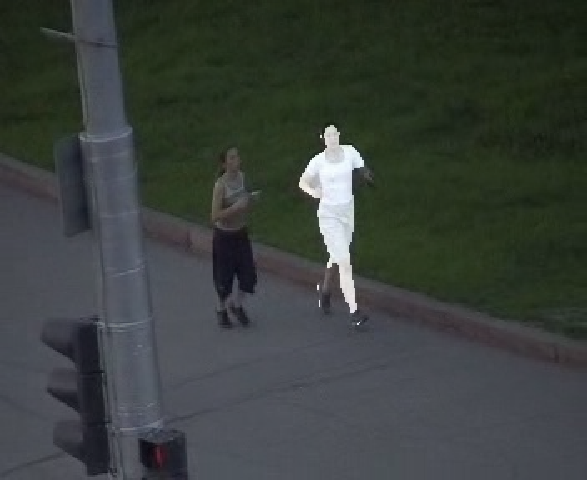}
\includegraphics[height=0.0798\textheight] {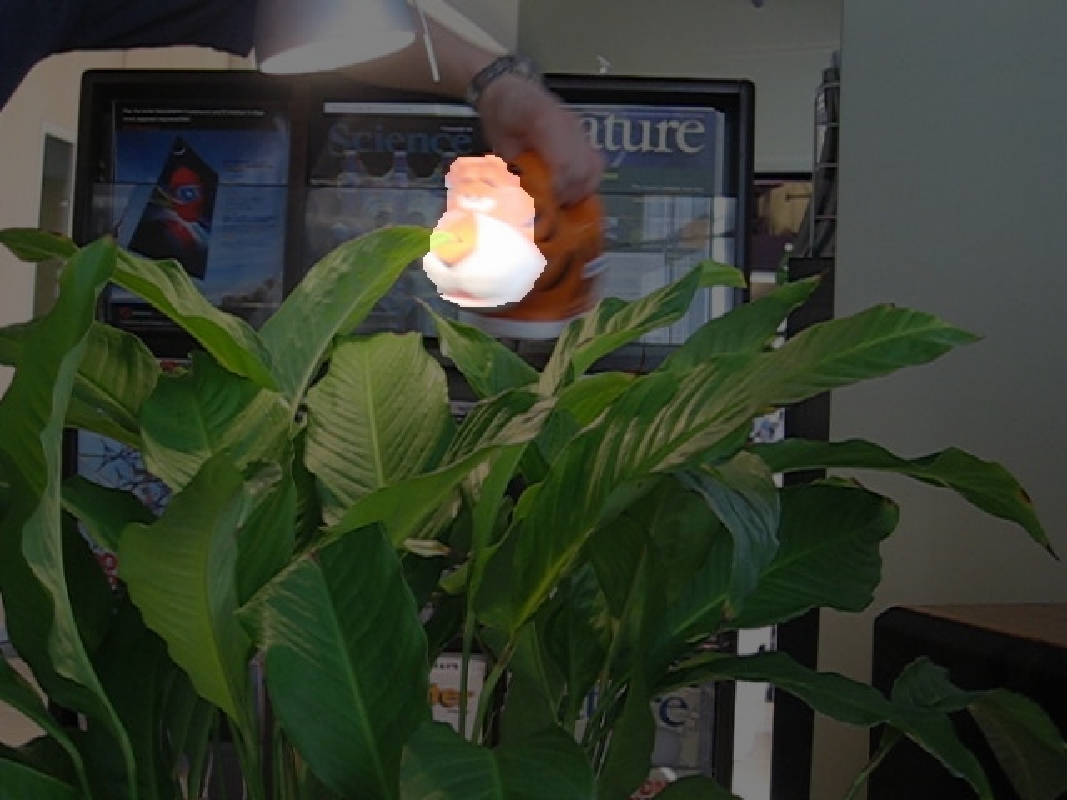}
\includegraphics[height=0.0798\textheight] {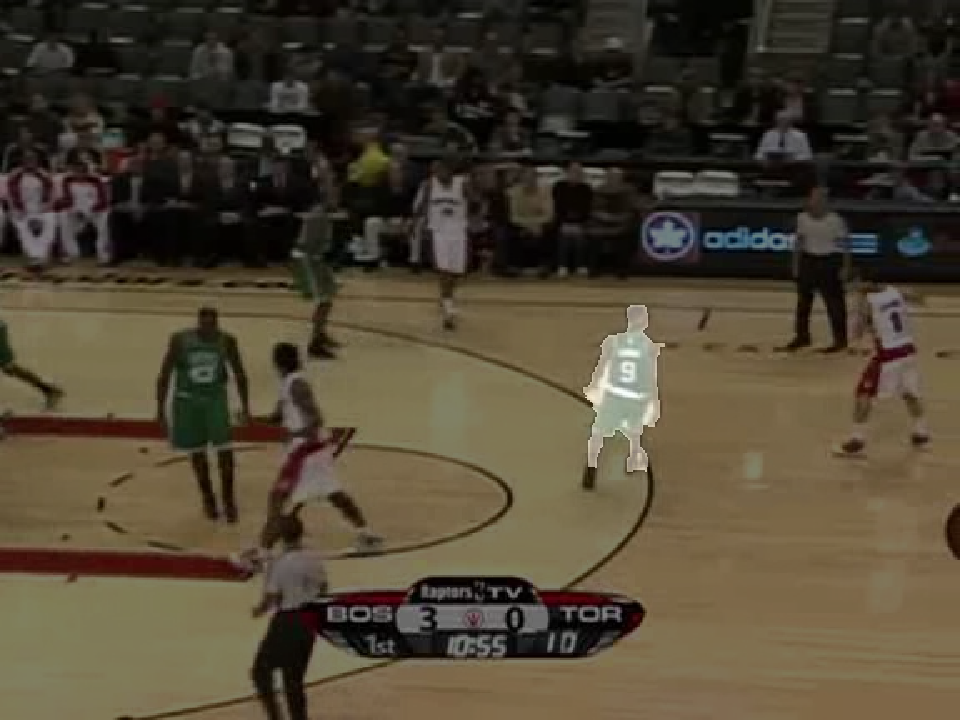}
\includegraphics[height=0.0798\textheight] {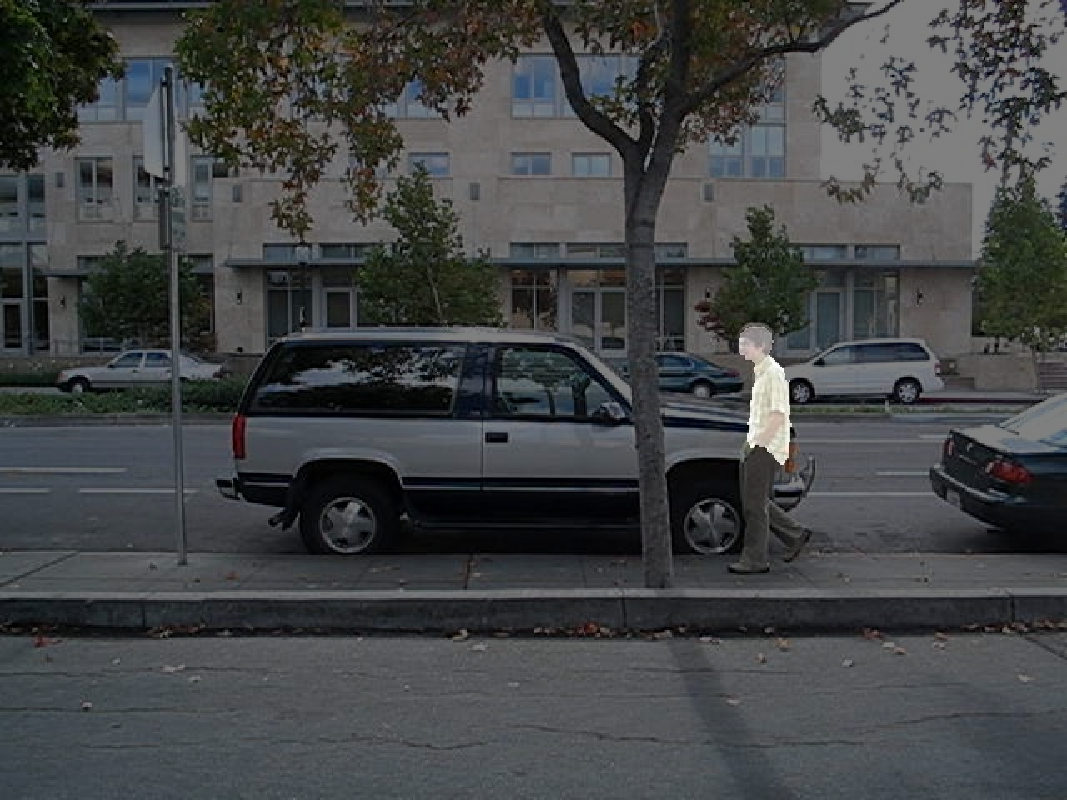}\\
\caption{Qualitative results for selected sequences: (from left to right) \textit{MotorRolling}, \textit{FaceOcc1}, \textit{Lemming},  \textit{Jogging}, \textit{Tiger}, \textit{Basketball} and \textit{David3}. (Row1) Comparisons to other trackers. (Row2) Target-specific saliency maps. (Row3) Segmentation by {\it GrabCut} with target-specific saliency maps.}
\label{fig:qualitative}
\end{minipage}
\end{figure*}

\paragraph{Quantitative results in bounding box} 

We evaluate our method quantitatively and make a comparative study with other methods in all the 50 benchmark sequences; the results are summarized in Figure~\ref{fig:quantitative_benchmark_OPE} for both of success rate and precision plots.
In both measures, our method outperforms all other trackers with substantial margins.
It is probably because the CNN features are more effective to represent high-level concept of target than hand-crafted ones although the network is trained offline for other purpose. 
We also compare our full algorithm with its reduced version denoted by $\text{Ours}_{\text{SVM}}$, which depends only on SVM scores as conventional tracking-by-detection algorithms do.
Our full algorithm achieves non-trivial performance improvement over the reduced version, which shows that our generative model based on target-specific saliency map is useful to localize target in general.
To gain more insight about the proposed algorithm, we evaluate the performance of trackers based on individual attributes provided in the benchmark dataset.
Note that the attributes describe 11 different types of tracking challenges and are annotated for each sequence.
Table~\ref{tab:att_success} and \ref{tab:att_precision} summarize the results in two different measures.
The numbers next to the attributes indicate the number of sequences involving the corresponding attribute.
As illustrated in the tables, our algorithm consistently outperforms other methods in almost all challenges, and our full algorithm is generally better than its reduced version.

%
\vspace{0.4cm}
\textbf{Quantitative results in segmentation~~} 
The proposed algorithm produces pixel-wise target segmentation using target-specific discriminative saliency map.
To evaluate segmentation accuracy, we select 9 video sequences from the online tracking benchmark dataset\footnote{Since accurate annotation of segmentation is labor intensive and time consuming, we selected a subset of sequences (typically short ones) for evaluation.} and annotate ground-truth segmentation for each sequence.
The selected sequences cover various attributes in tracking challenges, and the list of sequences with associated attributes are summarized in Table~\ref{tab:seq_attributes}. 

\begin{figure}[!t]
\centering
\subfigure{\includegraphics[width=0.8\linewidth]{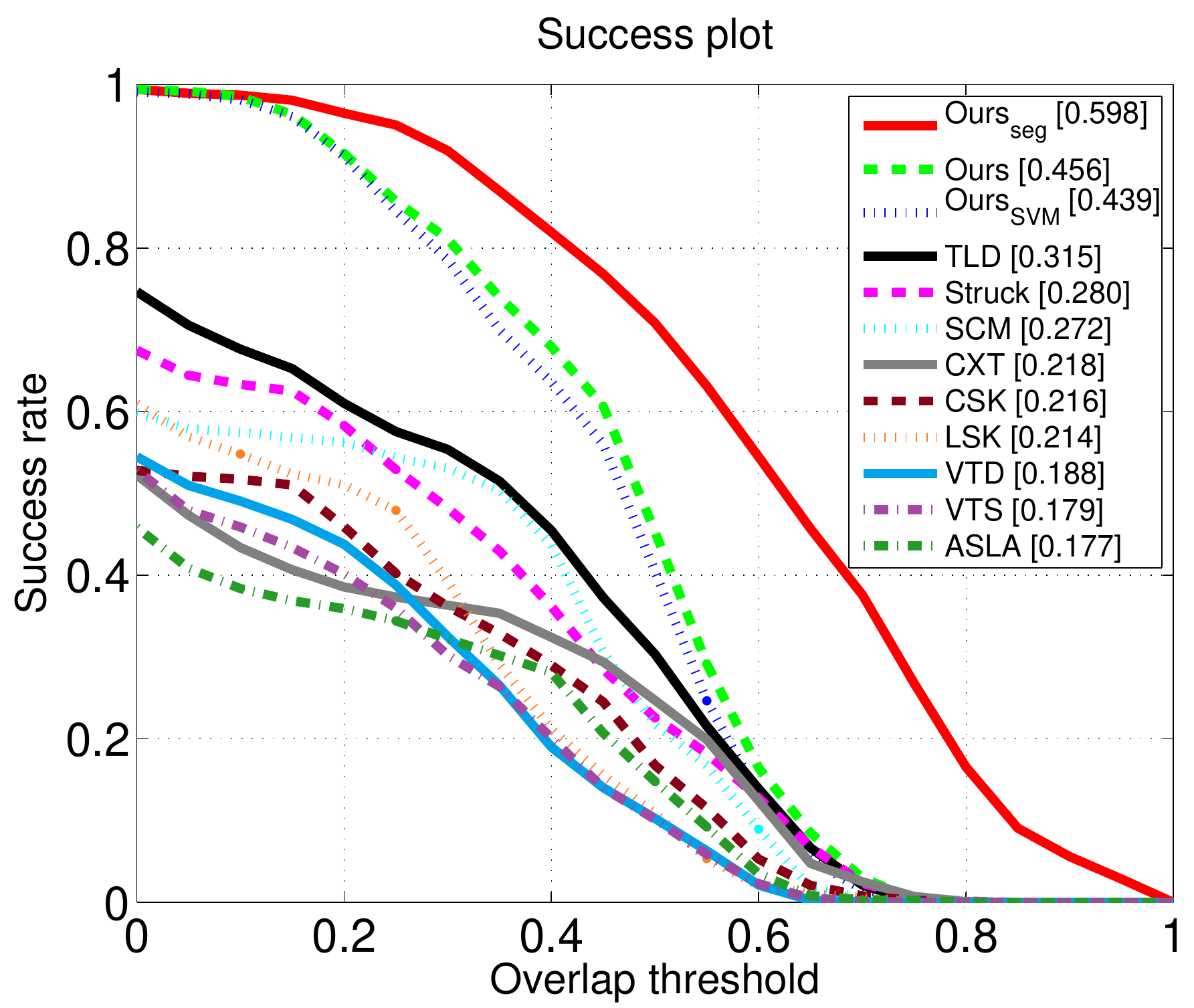}} \\
\caption{Average success plot over 9 selected sequences. Numbers in the legend indicate overall scores calculated by AUC.}
\label{fig:quantitative_segGT}
\end{figure}

The segmentation performance of the proposed algorithm is evaluated based on the overlap ratio---intersection over union---between ground-truth and identified target segmentation.
As other trackers used for comparison may not be able to generate pixel-wise segmentation, we employ their bounding box outputs as segmentation masks and compute the overlap ratio with respect to the ground-truth segmentation.
The results are presented by success plot as in Figure~\ref{fig:quantitative_segGT}, where $\text{Ours}_{\text{seg}}$ denotes the proposed algorithm with target segmentation.
According to Figure~\ref{fig:quantitative_segGT}, our method outperforms all other trackers with substantial margin.
Especially, we can observe a large performance improvement of the proposed target segmentation algorithm over our bonding box trackers denoted by Ours and $\text{Ours}_{\text{SVM}}$.
It suggests that the proposed target-specific saliency map is sufficiently accurate to estimate the target area in a video thus can be utilized to further improve tracking.

\begin{table}[!t] 
\begin{center}
\caption{List of sequences and their attributes used for segmentation performance evaluation. 
The set of sequences contains 10 attributes (out of 11 altogether) such as illumination variations (IV), out-of-plane rotation (OPR), scale variations (SV), occlusion (OCC), deformation (DEF), motion blur (MB), fast motion (FM), in-plane rotation (IPR), background clutter (BC) and low resolution (LR). 
The numbers in parentheses denote the number of frames.} 
\vskip 0.1in
\begin{tabular}{c|l}
Sequence name	&Attributes\\
\hline \hline
{\it Bolt} (350)			&OPR, OCC, DEF, IPR \\
{\it Coke} (291)			&IV, OPR, OCC, FM, IPR \\
{\it Couple} (140)			&OPR, SC, DEF	FM, BC \\
{\it Jogging} (307)		&OPR, OCC, DEF \\
{\it MotorRolling} (164)	&IV,	SC, MB, FM, IPR, BC, LR\\
{\it MountainBike} (228)	&OPR, IPR, BC \\
{\it Walking} (412)		&SC, OCC, DEF \\
{\it Walking2} (500)		&SC, OCC, LR \\
{\it Woman} (597)		&IV, OPR, SC, OCC, DEF, MB, FM\\
\hline 
\end{tabular}
\label{tab:seq_attributes}
\end{center}
\end{table}


\paragraph{Qualitative Results}
\label{sub:qualitative}
We present the results of several sequences in Figure~\ref{fig:qualitative}, where original frames with tracking results, target-specific saliency maps, and segmentation results are illustrated.
We can observe that our algorithm also demonstrates superior performance to other algorithms qualitatively.
\section{Conclusion}
\label{sec:conclusion}
We proposed a novel visual tracking algorithm based on pre-trained CNN, where
outputs from the last convolutional layer of the CNN are employed as generic feature descriptors of objects, and discriminative appearance models are learned online using an online SVM. 
With CNN features and learned discriminative model, we compute the target-specific saliency map by back-propagation, which highlights the discriminative target regions in spatial domain. 
Tracking is performed by sequential Bayesian filtering with the target-specific saliency map as observation.
The proposed algorithm achieves substantial performance gain over the existing state-of-the-art trackers and shows the capability for target segmentation.



\bibliographystyle{icml2015}
\bibliography{bib_tracking}

\begin{thebibliography}{43}
\providecommand{\natexlab}[1]{#1}
\providecommand{\url}[1]{\texttt{#1}}
\expandafter\ifx\csname urlstyle\endcsname\relax
  \providecommand{\doi}[1]{doi: #1}\else
  \providecommand{\doi}{doi: \begingroup \urlstyle{rm}\Url}\fi

\bibitem[Babenko et~al.(2011)Babenko, Yang, and Belongie]{babenko:11}
Babenko, Boris, Yang, Ming-Hsuan, and Belongie, Serge.
\newblock Robust object tracking with online multiple instance learning.
\newblock \emph{TPAMI}, 33, 2011.

\bibitem[Bao et~al.(2012)Bao, Wu, Ling, and Ji]{bao:12}
Bao, Chenglong, Wu, Yi, Ling, Haibin, and Ji, Hui.
\newblock Real time robust l1 tracker using accelerated proximal gradient
  approach.
\newblock In \emph{CVPR}, 2012.

\bibitem[Berg et~al.(2012)Berg, Deng, and Fei-Fei]{deng:12}
Berg, Alex, Deng, Jia, and Fei-Fei, L.
\newblock Large scale visual recognition challenge ({ILSVRC}).
\newblock \url{http://www.image-net.org/challenges/LSVRC/2012/}, 2012.

\bibitem[Bordes et~al.(2005)Bordes, Ertekin, Weston, and Bottou]{bordes:05}
Bordes, Antoine, Ertekin, Seyda, Weston, Jason, and Bottou, L\'{e}on.
\newblock Fast kernel classifiers with online and active learning.
\newblock \emph{JMLR}, 6, 2005.

\bibitem[Bordes et~al.(2007)Bordes, Bottou, Gallinari, and Weston]{bordes:07}
Bordes, Antoine, Bottou, L{\'e}on, Gallinari, Patrick, and Weston, Jason.
\newblock Solving multiclass support vector machines with larank.
\newblock In \emph{ICML}, 2007.

\bibitem[Cauwenberghs \& Poggio(2000)Cauwenberghs and Poggio]{cauwenberghs:01}
Cauwenberghs, Gert and Poggio, Tomaso.
\newblock Incremental and decremental support vector machine learning.
\newblock In \emph{NIPS}, 2000.

\bibitem[Diehl \& Cauwenberghs(2003)Diehl and Cauwenberghs]{diehl:03}
Diehl, C.P. and Cauwenberghs, G.
\newblock Svm incremental learning, adaptation and optimization.
\newblock In \emph{Proceedings of the International Joint Conference on Neural
  Networks}, 2003.

\bibitem[Dinh et~al.(2011)Dinh, Vo, and Medioni]{thang:11}
Dinh, Thang~Ba, Vo, Nam, and Medioni, G.
\newblock Context tracker: Exploring supporters and distracters in
  unconstrained environments.
\newblock In \emph{CVPR}, 2011.

\bibitem[Donahue et~al.(2014)Donahue, Jia, Vinyals, Hoffman, Zhang, Tzeng, and
  Darrell]{donahue:14}
Donahue, Jeff, Jia, Yangqing, Vinyals, Oriol, Hoffman, Judy, Zhang, Ning,
  Tzeng, Eric, and Darrell, Trevor.
\newblock Decaf: A deep convolutional activation feature for generic visual
  recognition.
\newblock In \emph{ICML}, 2014.

\bibitem[Fan et~al.(2010)Fan, Xu, Wu, and Gong]{fan:10}
Fan, Jialue, Xu, Wei, Wu, Ying, and Gong, Yihong.
\newblock Human tracking using convolutional neural networks.
\newblock \emph{Neural Networks}, 21, 2010.

\bibitem[Felzenszwalb et~al.(2010)Felzenszwalb, Girshick, McAllester, and
  Ramanan]{felzenszwalb:10}
Felzenszwalb, P.~F., Girshick, R.~B., McAllester, D., and Ramanan, D.
\newblock Object detection with discriminatively trained part-based models.
\newblock \emph{TPAMI}, 32, 2010.

\bibitem[Gall et~al.(2011)Gall, Yao, Razavi, Gool, and Lempitsky]{gall:11}
Gall, J., Yao, A., Razavi, N., Gool, L.~Van, and Lempitsky, V.
\newblock Hough forests for object detection, tracking, and action recognition.
\newblock \emph{TPAMI}, 33, 2011.

\bibitem[Girshick et~al.(2014)Girshick, Donahue, Darrell, and
  Malik]{girshick:14}
Girshick, Ross, Donahue, Jeff, Darrell, Trevor, and Malik, Jitendra.
\newblock Rich feature hierarchies for accurate object detection and semantic
  segmentation.
\newblock In \emph{CVPR}, 2014.

\bibitem[Grabner et~al.(2006)Grabner, Grabner, and Bischof]{grabner:06}
Grabner, H., Grabner, M., and Bischof, H.
\newblock Real-time tracking via on-line boosting.
\newblock In \emph{BMVC}, 2006.

\bibitem[Han et~al.(2008)Han, Comaniciu, Zhu, and Davis]{han:08}
Han, B., Comaniciu, D., Zhu, Y., and Davis, L.~S.
\newblock Sequential kernel density approximation and its application to
  real-time visual tracking.
\newblock \emph{TPAMI}, 30, 2008.

\bibitem[Hare et~al.(2011)Hare, Saffari, and Torr]{hare:11}
Hare, S., Saffari, A., and Torr, P. H~S.
\newblock Struck: Structured output tracking with kernels.
\newblock In \emph{ICCV}, 2011.

\bibitem[Hariharan et~al.(2014)Hariharan, Arbel\'{a}ez, Girshick, and
  Malik]{hariharan:14}
Hariharan, Bharath, Arbel\'{a}ez, Pablo, Girshick, Ross, and Malik, Jitendra.
\newblock Simultaneous detection and segmentation.
\newblock In \emph{ECCV}, 2014.

\bibitem[He et~al.(2014)He, Zhang, Ren, and Sun]{kaiming:14}
He, Kaiming, Zhang, Xiangyu, Ren, Shaoqing, and Sun, Jian.
\newblock Spatial pyramid pooling in deep convolutional networks for visual
  recognition.
\newblock In \emph{ECCV}, 2014.

\bibitem[Henriques et~al.(2012)Henriques, Caseiro, Martins, and
  Batista]{henriques:12}
Henriques, Joao~F., Caseiro, Rui, Martins, Pedro, and Batista, Jorge.
\newblock Exploiting the circulant structure of tracking-by-detection with
  kernels.
\newblock In \emph{ECCV}, 2012.

\bibitem[Jia et~al.(2012)Jia, Lu, and Yang]{jia:12}
Jia, Xu, Lu, Huchuan, and Yang, Ming-Hsuan.
\newblock Visual tracking via adaptive structural local sparse appearance
  model.
\newblock In \emph{CVPR}, 2012.

\bibitem[Jia(2013)]{jia:13}
Jia, Y.
\newblock {Caffe}: An open source convolutional architecture for fast feature
  embedding.
\newblock {\url http://caffe.berkeleyvision.org/}, 2013.

\bibitem[Kalal et~al.(2012)Kalal, Mikolajczyk, and Matas]{kalal:12}
Kalal, Zdenek, Mikolajczyk, Krystian, and Matas, Jiri.
\newblock Tracking-{L}earning-{D}etection.
\newblock \emph{TPAMI}, 2012.

\bibitem[Karayev et~al.(2014)Karayev, Trentacoste, Han, Agarwala, Darrell,
  Hertzmann, and Winnemoeller]{karayev:14}
Karayev, Sergey, Trentacoste, Matthew, Han, Helen, Agarwala, Aseem, Darrell,
  Trevor, Hertzmann, Aaron, and Winnemoeller, Holger.
\newblock Recognizing image style.
\newblock In \emph{BMVC}, 2014.

\bibitem[Krizhevsky et~al.(2012)Krizhevsky, Sutskever, and
  Hinton]{krizhevsky:12}
Krizhevsky, A., Sutskever, I., and Hinton, G.~E.
\newblock {ImageNet Classification with Deep Convolutional Neural Networks}.
\newblock In \emph{NIPS}, 2012.

\bibitem[Kwon \& Lee(2010)Kwon and Lee]{kwon:09}
Kwon, Junseok and Lee, Kyoung-Mu.
\newblock Visual tracking decomposition.
\newblock In \emph{CVPR}, 2010.

\bibitem[Kwon \& Lee(2011)Kwon and Lee]{kwon:11}
Kwon, Junseok and Lee, Kyoung~Mu.
\newblock Tracking by sampling trackers.
\newblock In \emph{ICCV}, 2011.

\bibitem[Li et~al.(2014)Li, Li, and Porikli]{li:14}
Li, H., Li, Y., and Porikli, F.
\newblock Deeptrack: Learning discriminative feature representations by
  convolutional neural networks for visual tracking.
\newblock In \emph{BMVC}, 2014.

\bibitem[Liu et~al.(2011)Liu, Huang, Yang, and Kulikowski]{liu:11}
Liu, Baiyang, Huang, Junzhou, Yang, Lin, and Kulikowski, Casimir~A.
\newblock Robust tracking using local sparse appearance model and k-selection.
\newblock In \emph{CVPR}, 2011.

\bibitem[Mei \& Ling(2009)Mei and Ling]{mei:09}
Mei, Xue and Ling, Haibin.
\newblock Robust visual tracking using $l$1 minimization.
\newblock In \emph{ICCV}, 2009.

\bibitem[Oquab et~al.(2014)Oquab, Bottou, Laptev, and Sivic]{oquab:14}
Oquab, M., Bottou, L., Laptev, I., and Sivic, J.
\newblock Learning and transferring mid-level image representations using
  convolutional neural networks.
\newblock In \emph{CVPR}, 2014.

\bibitem[Ross et~al.(2004)Ross, Lim, and Yang]{ross:04}
Ross, D., Lim, J., and Yang, M.-H.
\newblock Adaptive probabilistic visual tracking with incremental subspace
  update.
\newblock In \emph{ECCV}, 2004.

\bibitem[Rother et~al.(2004)Rother, Kolmogorov, and Blake]{rother:04}
Rother, Carsten, Kolmogorov, Vladimir, and Blake, Andrew.
\newblock "grabcut": Interactive foreground extraction using iterated graph
  cuts.
\newblock In \emph{SIGGRAPH}, 2004.

\bibitem[Saffari et~al.(2010)Saffari, Godec, Pock, Leistner, and
  Bischof]{saffari:10}
Saffari, A., Godec, M., Pock, T., Leistner, C., and Bischof, H.
\newblock Online multi-class lpboost.
\newblock In \emph{CVPR}, 2010.

\bibitem[Schulter et~al.(2011)Schulter, Leistner, Roth, Gool, , and
  Bischof]{schulter:11}
Schulter, Samuel, Leistner, Christian, Roth, Peter~M., Gool, Luc~Van, , and
  Bischof, Horst.
\newblock Online hough-forests.
\newblock In \emph{BMVC}, 2011.

\bibitem[Sermanet et~al.(2014)Sermanet, Eigen, Zhang, Mathieu, Fergus, and
  LeCun]{sermanet:14}
Sermanet, Pierre, Eigen, David, Zhang, Xiang, Mathieu, Michael, Fergus, Rob,
  and LeCun, Yann.
\newblock Overfeat: Integrated recognition, localization and detection using
  convolutional networks.
\newblock In \emph{ICLR}, 2014.

\bibitem[Sevilla-Lara \& Learned-Miller(2012)Sevilla-Lara and
  Learned-Miller]{sevilla:12}
Sevilla-Lara, L. and Learned-Miller, E.
\newblock Distribution fields for tracking.
\newblock In \emph{CVPR}, 2012.

\bibitem[Simonyan et~al.(2014)Simonyan, Vedaldi, and Zisserman]{simonyan:14}
Simonyan, Karen, Vedaldi, Andrea, and Zisserman, Andrew.
\newblock Deep inside convolutional networks: Visualising image classification
  models and saliency maps.
\newblock In \emph{ICLR Workshop}, 2014.

\bibitem[Toshev \& Szegedy(2014)Toshev and Szegedy]{Toshev:14}
Toshev, A. and Szegedy, C.
\newblock Deeppose: Human pose estimation via deep neural networks.
\newblock In \emph{CVPR}, 2014.

\bibitem[Wang \& Yeung(2013)Wang and Yeung]{wang:13}
Wang, Naiyan and Yeung, Dit-Yan.
\newblock Learning a deep compact image representation for visual tracking.
\newblock In \emph{NIPS}, 2013.

\bibitem[Wu et~al.(2013)Wu, Lim, and Yang]{wu:13}
Wu, Yi, Lim, Jongwoo, and Yang, Ming-Hsuan.
\newblock Online object tracking: A benchmark.
\newblock In \emph{CVPR}, 2013.

\bibitem[Zhang et~al.(2014)Zhang, Donahue, Girshick, and Darrell]{zhang:14}
Zhang, Ning, Donahue, Jeff, Girshick, Ross, and Darrell, Trevor.
\newblock Part-based {R-CNNs} for fine-grained category detection.
\newblock In \emph{ECCV}, 2014.

\bibitem[Zhang et~al.(2012)Zhang, Ghanem, Liu, and Ahuja]{zhang:12}
Zhang, Tianzhu, Ghanem, Bernard, Liu, Si, and Ahuja, Narendra.
\newblock Robust visual tracking via multi-task sparse learning.
\newblock In \emph{CVPR}, 2012.

\bibitem[Zhong et~al.(2012)Zhong, Lu, and Yang]{zhong:12}
Zhong, Wei, Lu, Huchuan, and Yang, Ming-Hsuan.
\newblock Robust object tracking via sparsity-based collaborative model.
\newblock In \emph{CVPR}, 2012.

\end{thebibliography}

\end{document}